\documentclass[letterpaper]{article} % DO NOT CHANGE THIS
\usepackage{aaai24}  % DO NOT CHANGE THIS
\usepackage{times}  % DO NOT CHANGE THIS
\usepackage{helvet}  % DO NOT CHANGE THIS
\usepackage{courier}  % DO NOT CHANGE THIS
\usepackage[hyphens]{url}  % DO NOT CHANGE THIS
\usepackage{graphicx} % DO NOT CHANGE THIS
\usepackage{natbib}  % DO NOT CHANGE THIS AND DO NOT ADD ANY OPTIONS TO IT
\usepackage{caption} % DO NOT CHANGE THIS AND DO NOT ADD ANY OPTIONS TO IT
\usepackage{algorithm}
\usepackage{amsmath}
\usepackage[ruled,boxed,algo2e]{algorithm2e}
\usepackage{array}
\usepackage{multirow}
\usepackage{ragged2e}
\usepackage{adjustbox}
\usepackage{tabularray}
\usepackage{amssymb}
\usepackage{newfloat}
\usepackage{listings}
\DeclareCaptionStyle{ruled}{labelfont=normalfont,labelsep=colon,strut=off} % DO NOT CHANGE THIS
\floatstyle{ruled}
\newfloat{listing}{tb}{lst}{}
\floatname{listing}{Listing}
\title{MWSIS: Multimodal Weakly Supervised Instance Segmentation with 2D Box Annotations for Autonomous Driving}
\author{
    Guangfeng Jiang\textsuperscript{\rm 1},
    Jun Liu\textsuperscript{\rm 1}\thanks{Corresponding authors.},
    Yuzhi Wu\textsuperscript{\rm 1},
    Wenlong Liao\textsuperscript{\rm 2},
    Tao He\textsuperscript{\rm 3},
    Pai Peng\textsuperscript{\rm 3}\footnotemark[1]
}
\affiliations{
    \textsuperscript{\rm 1}Department of Electronic Engineering and Information Science, University of Science and Technology of China\\
    \textsuperscript{\rm 2}Shanghai Jiao Tong University\\
    \textsuperscript{\rm 3}COWAROBOT\\
    jgf1998@mail.ustc.edu.cn, junliu@ustc.edu.cn,
    pengpai@cowarobot.com
}
\begin{document}
\maketitle
\begin{abstract}
    Instance segmentation is a fundamental research in computer vision, especially in autonomous driving. However, manual mask annotation for instance segmentation is quite time-consuming and costly. To address this problem, some prior works attempt to apply weakly supervised manner by exploring 2D or 3D boxes. However, no one has ever successfully segmented 2D and 3D instances simultaneously by only using 2D box annotations, which could further reduce the annotation cost by an order of magnitude. Thus, we propose a novel framework called Multimodal Weakly Supervised Instance Segmentation (MWSIS), which incorporates various fine-grained label correction modules for both 2D and 3D modalities, along with a new multimodal cross-supervision approach. In the 2D pseudo label generation branch, the Instance-based Pseudo Mask Generation (IPG) module utilizes predictions for self-supervised correction. Similarly, in the 3D pseudo label generation branch, the Spatial-based Pseudo Label Generation (SPG) module generates pseudo labels by incorporating the spatial prior information of the point cloud. To further refine the generated pseudo labels, the Point-based Voting Label Correction (PVC) module utilizes historical predictions for correction. Additionally, a Ring Segment-based Label Correction (RSC) module is proposed to refine the predictions by leveraging the depth prior information from the point cloud. Finally, the Consistency Sparse Cross-modal Supervision (CSCS) module reduces the inconsistency of multimodal predictions by response distillation. Particularly, transferring the 3D backbone to downstream tasks not only improves the performance of the 3D detectors, but also outperforms fully supervised instance segmentation with only 5\% fully supervised annotations. On the Waymo dataset, the proposed framework demonstrates significant improvements over the baseline, especially achieving 2.59\% mAP and 12.75\% mAP increases for 2D and 3D instance segmentation tasks, respectively. The code is available at https://github.com/jiangxb98/mwsis-plugin.
\end{abstract}

\section{Introduction}
    In the field of autonomous driving, instance segmentation of images and point clouds are two important research directions, each providing fine-grained perception results in their respective modalities. However, the fully supervised instance segmentation approaches \cite{He2017MaskRCNN, tian2020Condinst, Cheng2021Maskformer, Jiang2020PointGroupDP, Wu20223DIA} rely on pixel-wise or point-wise instance annotations, which are labor-intensive and costly \cite{Behley2019ADF, cheng2022pointly}. As a result, many weakly supervised approaches \cite{hsu2019BBTP, lee2021bbam, shen2021parallel, Tian2020BoxInstHI, wang2021BoxCaseg, cheng2022pointly, Chibane2022Box2MaskWS, dong2022rwseg} using weak labels have emerged, e.g., BoxInst \cite{Tian2020BoxInstHI} is trained with 2D box annotations to predict 2D masks in the image space, and Box2Mask\cite{Chibane2022Box2MaskWS} is trained with 3D box annotations to predict 3D masks in the point cloud space. However, the 3D box annotation is also costly.
    
    These early weakly supervised works are mostly limited to a single modality. Thanks to the development of multimodal datasets, e.g., nuScenes \cite{Caesar2019nuScenesAM} and Waymo \cite{Sun2019waymo}, which offer millions of images and hundreds of thousands of LiDAR scans. To the best of our knowledge, LWSIS \cite{li2023lwsis} is the first to focus on multimodal instance segmentation, which employs LiDAR points inside 3D box annotations to guide the 2D instance segmentation. However, LWSIS only utilizes multimodal information to obtain a single-modal segmentor and uses costly 3D box annotations. This leads us to the question: \textbf{Is it possible to leverage well synchronized \& aligned camera and LiDAR data, along with only much simpler and more cost-effective 2D box annotations, to train both image and point cloud segmentors simultaneously via weak supervision?} 
    
    Motivated by this question, we propose a novel framework called Multimodal Weakly Supervised Instance Segmentation (MWSIS). It trains both 2D and 3D segmentors using only 2D box annotations, leveraging complementary information from different modalities through cross-modal distillation.

    However, our framework inevitably faces two challenges: one is how to deal with the low signal-to-noise ratio introduced by 2D box annotations as weak supervision signals, and the other is how to implement cross-modal distillation. To address these challenges, we devise several modules for pseudo label generation and correction at various granularity, such as the Instance-based Pseudo Mask Generation (IPG) module, the Spatial-based Pseudo Label Generation (SPG) module, the Point-based Voting Label Correction (PVC) module, and the Ring Segment-based Label Correction (RSC) module, to improve the quality of pseudo labels and filter out noise. Additionally, we design a Consistency Sparse Cross-modal Supervision (CSCS) module to achieve cross-modal supervision. The detailed implementation of the above modules is described in Sec. \ref{sec: method}.
    
    In short, our contributions are summarized in four folds:
    \begin{itemize}
        \item To the best of our knowledge, we are the first to use the 2D box annotations as the sole external supervision signal to train both image and point cloud instance segmentors simultaneously.
        \item We propose various fine-grained label correction modules for different modalities, including instance-based, spatial-based, point-based, and ring segment-based modules. These modules not only enhance the instance segmentation performance, but also improve the quality of the pseudo label.
        \item We propose a novel cross-modal supervision method, named CSCS, which exploits the complementary properties of the point cloud and image modalities. This method improves the performance of the segmentors.
        \item Our framework can be used as a pre-training method to improve the performance of 3D downstream tasks such as semantic segmentation, instance segmentation, and object detection.
    \end{itemize}

\section{Related Work}
    The development of multimodal datasets, such as KITTI \cite{kitti}, nuScenes \cite{Caesar2019nuScenesAM}, and Waymo \cite{Sun2019waymo}, has greatly contributed to the progress of weakly supervised methods in various tasks, including semantic segmentation \cite{Wei2020MultiPathRM, Hu2021SQNWS, Unal2022ScribbleSupervisedLS}, instance segmentation \cite{Chibane2022Box2MaskWS,li2023lwsis}, and object detection \cite{Qin2020vs3d, Wei2021FGR, Peng2022WeakM3DTW, Liu2022MAPGenAA, Liu2022MultimodalTF}. The weakly supervised instance segmentation task is to extract objects from an image or point cloud using simpler annotations than mask annotations. These simpler annotations include image-level labels \cite{Cholakkal2019ObjectCA, Ge2019LabelPEnetSL, shen2021parallel}, points \cite{lee2021weakly,cheng2022pointly,li2023lwsis}, scribble \cite{Li2023ScribbleVCSM, Chen2023ScribbleSegSI}, and boxes \cite{hsu2019BBTP, lee2021bbam, wang2021BoxCaseg, Chibane2022Box2MaskWS}.

\subsection{2D Weakly Supervised Instance Segmentation}
    Among models trained with image-level supervision, PDSL \cite{shen2021parallel} adopts self-supervised learning to learn class-independent foreground segmentation. Among works using box-level supervision, BBTP \cite{hsu2019BBTP} is the first one that uses 2D boxes to generate instance masks in the formulation of multiple instance learning, leveraging the tightness of boxes to predict instance masks. BBAM \cite{lee2021bbam} utilizes the semantics extracted by a trained detector to generate pseudo labels, which are then used to train semantic segmentation and instance segmentation networks. BoxInst \cite{Tian2020BoxInstHI} makes use of the local color consistency constraints and designs projection loss and pairwise similarity loss to supervise the mask branch of CondInst \cite{tian2020Condinst}. Among point-level supervised methods, PointSup \cite{cheng2022pointly} adds random sampling points as segmentation annotations based on 2D box annotations. LWSIS \cite{li2023lwsis} is the first work to incorporate LiDAR points, which utilizes more precise guidance of points inside the 3D box as a supervision signal for 2D instance segmentation, and has achieved significant improvement without introducing additional network parameters.
    
\subsection{3D Weakly Supervised Instance Segmentation}
    There is only a small amount of work on the weakly supervised 3D instance segmentation. As one of the methods using 3D box annotations, Box2Mask \cite{Chibane2022Box2MaskWS} is inspired by classical Hough voting, in which each point directly votes for a 3D box, and it uses the IoU-guided NMC method to obtain the clustered box, which is then back-project to the point cloud to obtain the instance segmentation result. RWSeg \cite{dong2022rwseg} employs self-attention and random walk to propagate semantic and instance information to unknown regions, respectively.

    Basically, these methods mentioned above can be divided into two categories: single-modal weakly supervised models like \cite{Tian2020BoxInstHI, lee2021bbam, Chibane2022Box2MaskWS, dong2022rwseg} and multimodal weakly supervised models for a single-modal task like \cite{li2023lwsis}. The aforementioned single-modal approaches utilize the information only from a single modality and do not take advantage of complementary information from multiple modalities, while the multimodal approaches leverage information from other modalities to acquire single-modal segmentors. In this paper, we examine the first attempt at multimodal weakly supervised instance segmentation to acquire the multimodal segmentors simultaneously, while using only 2D box annotations.
% framework pipeline
\begin{figure*}[t]
\begin{center}
   \includegraphics[width=1.0\linewidth]{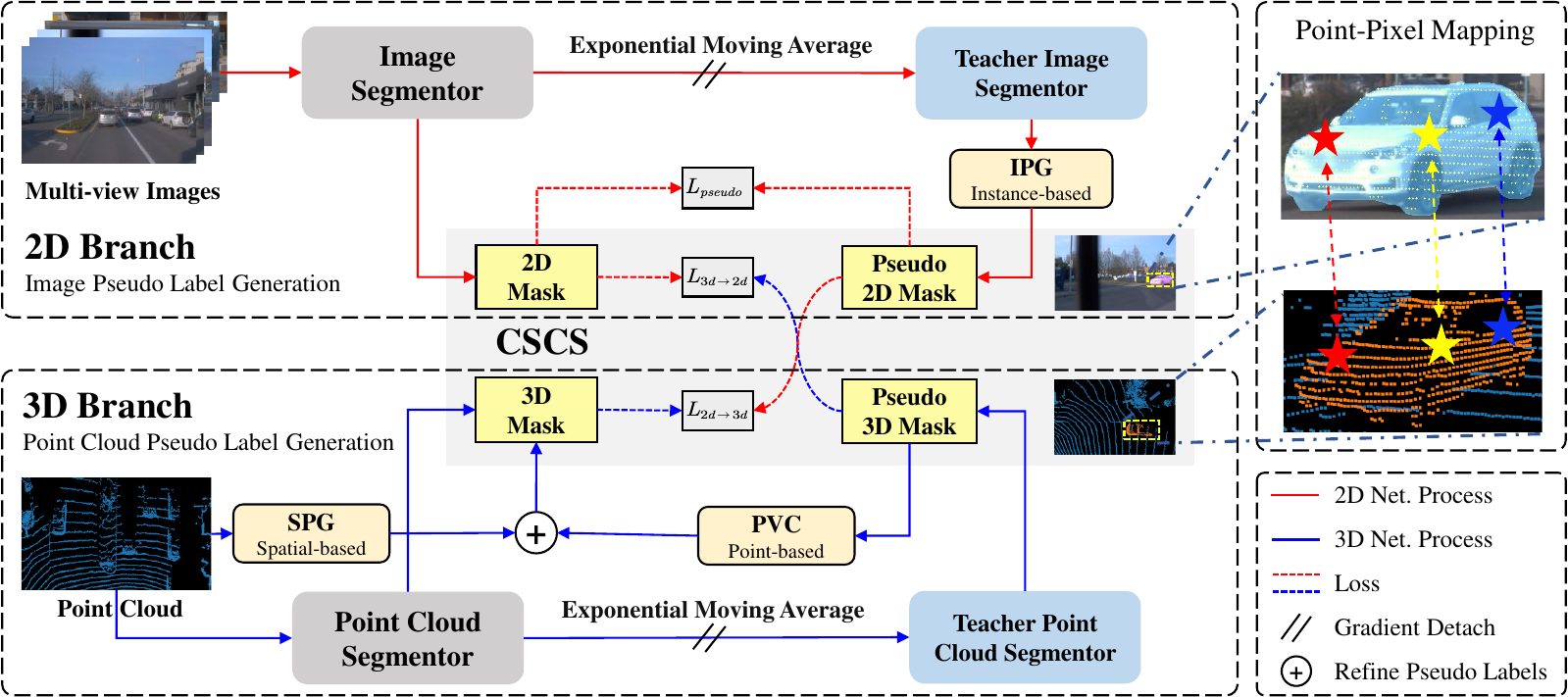}
\end{center}
   \caption{Overview of MWSIS framework. The framework mainly consists of three parts: image pseudo label generation branch, point cloud pseudo label generation branch, and CSCS module. In the 2D branch, the pseudo 2D masks generated by the IPG module self-supervise the 2D masks. In the 3D branch, the label is refined by the SPG module and PVC module to supervise the 3D Masks. Finally, the pseudo masks from the teacher model are used for cross-supervision in the CSCS module.}
\label{fig:architecture}
\end{figure*}

\section{Method}
\label{sec: method}
    In this section, we provide the implementation details of our MWSIS framework. The framework consists of two separate modal branches and a multimodal cross-supervision module. Each branch includes a trainable student segmentor and a teacher segmentor updated by EMA (Exponential Moving Average). The cross-supervision module utilizes the output of the teacher segmentor to perform cross-supervision on the student segmentor of another modality through response distillation. As shown in Fig. \ref{fig:architecture}, in the image branch (top), we choose the BoxInst as the baseline and employ the instance-based IPG module for self-supervision (Sec. \ref{Instance-based Pseduo Mask Generation}). In the point cloud branch (bottom), we employ a plain voxel-based SparseUNet \cite{shi2020parta2} segmentor as the baseline, and we introduce the SPG module based on the spatial priors of point clouds, such as depth and Euclidean distance, to generate high-quality pseudo instance masks for the point cloud from 2D box annotations. We also present the point-based PVC module based on voting using historical results from the teacher segmentor to refine the pseudo labels, and the ring segment-based RSC module to refine the student segmentor predictions (Sec. \ref{3D label correction}). Finally, the CSCS module incorporates the mean teacher \cite{Tarvainen2017MeanTeacher} and CPS \cite{Chen2021CPS} principles and employs response distillation to ensure consistency between multimodal masks (Sec. \ref{CSCS}).
\subsection{Preliminary}
\label{Point cloud projection}
   We define the input multimodal data as $\left\{\mathbf{P}, \mathbf{I}\right\}$, where $\mathbf{P}\in \mathbb{R}^{N_{in}\times C_{in}}$ denotes $N_{in}$ points with $C_{in} \text{-dimensional}$ input features (e.g., 3D cordinates and reflectance) and $\mathbf{I} \in \mathbb{R}^{N_{cam}\times 3\times H_{in} \times W_{in}}$ denotes the multi-view RGB images of dimension $3\times H_{in}\times W_{in}$ obtained from $N_{cam}$ cameras. Using sensor calibration of the dataset, we can project 3D points onto 2D images, and obtain the point-pixel mapping relationship between spatial points $\mathbf{P}_{3d}\in\mathbb{R}^{N_{in}\times 3}$ and pixel points $\mathbf{P}_{2d}\in\mathbb{R}^{N_{in}\times 2}$.
    
\subsection{2D -- Image Pseudo Label Generation}
    \label{Instance-based Pseduo Mask Generation}
    % ipg figure
    \begin{figure}[t]
    \begin{center}
       \includegraphics[width=1.0\linewidth]{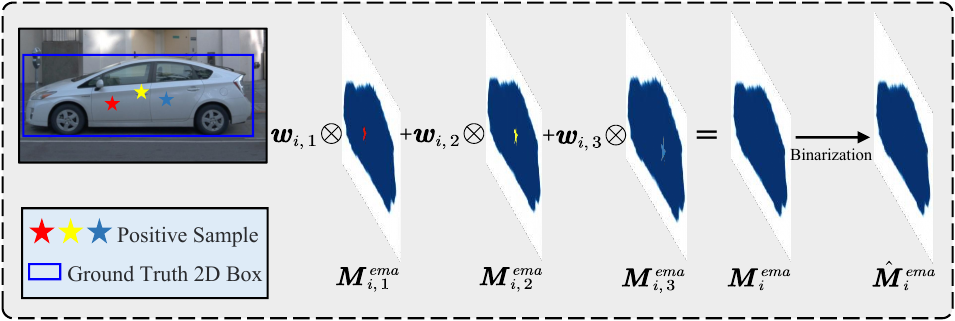}
    \end{center}
       \caption{IPG pseudo mask generation process.}
    \label{fig:ipg}
    \end{figure}
    
    \textbf{Instance-based Pseudo Mask Generation (IPG)}. We chose BoxInst \cite{Tian2020BoxInstHI} as our 2D baseline. In anchor-free methods \cite{tian2020Condinst,tian2019fcos}, multiple positive samples can be assigned to the same ground truth (GT) instance during the assignment, resulting in inconsistency predictions, as shown by the different colors of stars in Fig. \ref{fig:ipg}. The masks with higher confidence scores and closer to the GT boxes center are considered to be more reliable. In this view, we propose the instance-based IPG module, as illustrated in Fig. \ref{fig:ipg}. Specifically, we calculate the IoUs between the predicted boxes and the GT boxes and weight the corresponding predicted masks based on the IoUs and scores. The formula is defined as:
    
    \begin{equation}
    \label{eq:mdw}
    \begin{aligned}
        \mathbf{M}_{i}^{ema} &= \sum_{j}^{N_i}w_{i,j}\mathbf{M}_{i,j}^{ema} \\
        w_{i,j} &= \frac{s_{i,j}e^{k IoU_{i,j}}}{\sum_j^{N_i}s_{i,j}e^{k IoU_{i,j}}}
    \end{aligned}
    \end{equation}
    where $\mathbf{M}_i^{ema}$ is the predicted probability map corresponding to the $i$-th GT box, $N_i$ is the number of positive samples assigned to the $i$-th GT box, $\mathbf{M}_{i,j}^{ema}$ is the $j$-th mask corresponding to the $i$-th GT box, $k$ is a hyperparameter, $s_{i,j}$ and $IoU_{i,j}$ are the confidence score and IoU of the $j$-th predicted box to the $i$-th GT box, respectively.
    
    After obtaining the weighted probability map, we set two thresholds $\tau_{low}$ and $\tau_{high}$ to obtain the pseudo mask $\hat{\mathbf{M}}^{ema}_i$. Then we apply the pseudo mask $\hat{\mathbf{M}}^{ema}_i$ as the self-supervision signal. The self-supervision loss function is defined as:
    \begin{equation}
    \label{eq:pseudo loss}
    \begin{aligned}
        L_{pseudo} = L_{pseudo}\left(\mathbf{M}_{pred},\hat{\mathbf{M}}^{ema}\right)\\
    \end{aligned}
    \end{equation}
    where $L$ consists of two terms: binary cross-entropy loss $L_{bce}$ and dice loss $L_{dice}$, $\mathbf{M}_{pred}$ is the predicted masks.
   
\subsection{3D -- Point Cloud Pseudo Label Generation}
\label{3D label correction}
\begin{figure}[t]
    \begin{center}
       \includegraphics[width=1.0\linewidth]{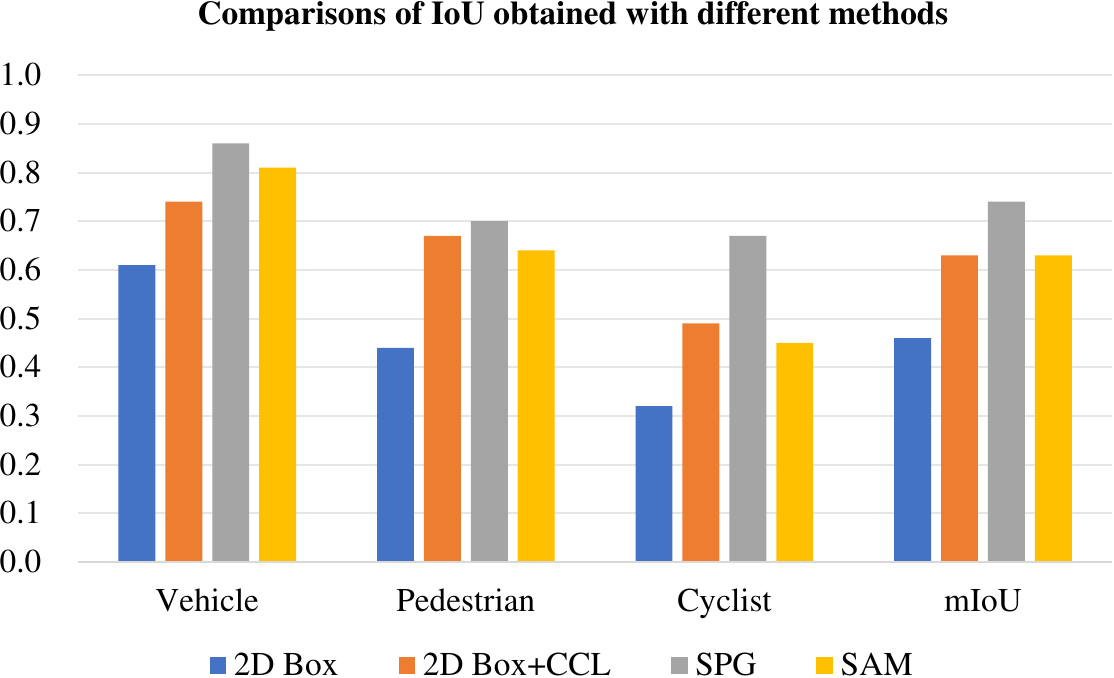}
    \end{center}
       \caption{Comparisons of IoU obtained with different methods on Waymo validation dataset. SAM means the process of obtaining masks through the use of SAM \cite{Kirillov2023SAM}, where 2D boxes are employed as prompts.}
    \label{fig:comparison of mIoU results}
    \end{figure}

\begin{figure}[t]
    \begin{center}
       \includegraphics[width=1.0\linewidth]{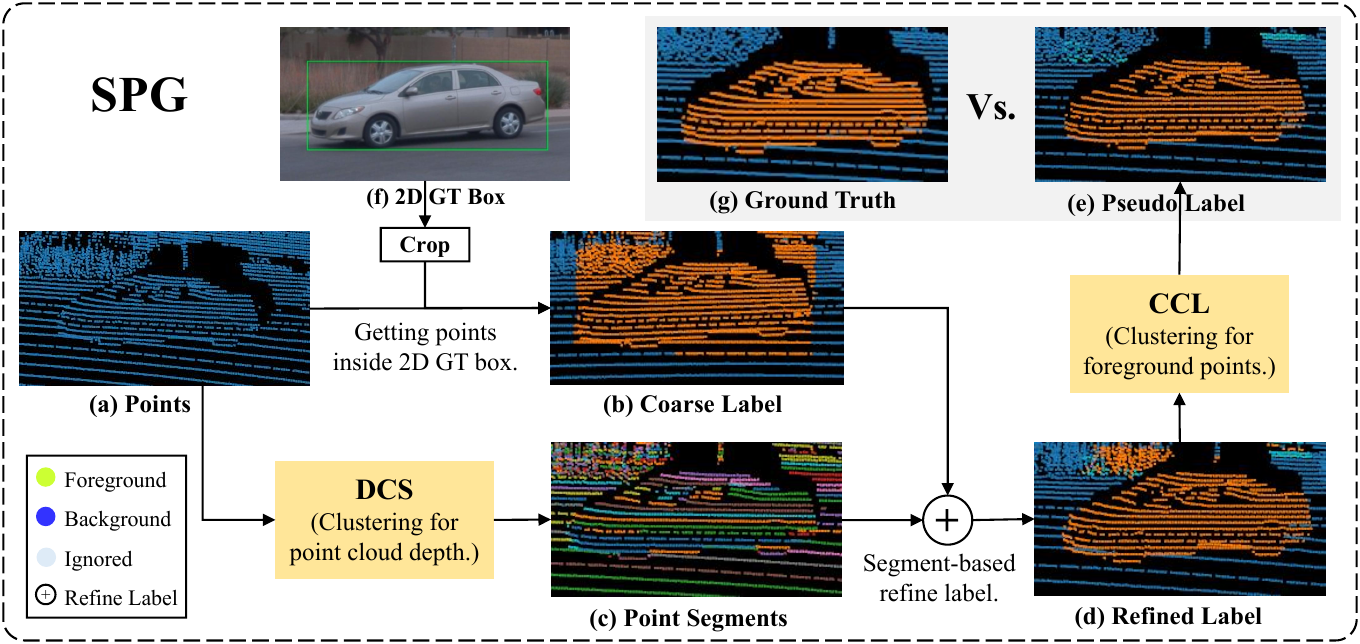}
    \end{center}
       \caption{The overview of SPG.}
    \label{fig:plg2}
    \end{figure}

    The input point cloud $\mathbf{P}$ can be roughly partitioned into potential foreground points $\mathbf{P}_{in}$ and background points $\mathbf{P}_{out}$ by 3D frustums bounded by 2D boxes. However, as shown in Fig. \ref{fig:comparison of mIoU results} and Fig. \ref{fig:plg2}(b), the foreground partition still contains a large number of background points, which is suboptimal to be treated as the pseudo label directly. Thus, we adopt a simple and efficient CCL \cite{ccl} clustering algorithm to partition the potential foreground points into clusters and select the largest connected component as the foreground points corresponding to the instance. To further improve the quality of the pseudo instance masks, we propose three label correction modules: the SPG, the PVC, and the RSC.

    \noindent\textbf{Spatial-based Pseudo Label Generation (SPG).} We notice that LiDAR points obtained from each laser beam exhibit a certain pattern. If the depth changes suddenly, it is likely that a different object is scanned. In Fig. \ref{fig:plg2}(c), points of the same color denote smooth depth variations, such as ground points, while points of different colors denote non-smooth depth variations. To leverage the prior information about the depth variation of the point cloud, we propose the Depth Clustering Segment (DCS) algorithm whose details are given in the following.
    
    % plg2
    
    A LiDAR sensor returns an $M \times N$ measurement matrix in a single scan, where $M$ denotes the number of beams and $N$ denotes the number of measurements. The idea of the DCS algorithm is to traverse each row and record the depth of each column. If the depth changes smoothly, the current point belongs to the same ring segment. Otherwise, it belongs to a different ring segment. The algorithm is shown in Appendix A.1.
    
    Fig. \ref{fig:plg2} shows the process of generating 3D pseudo labels by the SPG. Firstly, points are cropped by 2D GT boxes to obtain coarse pseudo labels in Fig. \ref{fig:plg2}(b), where the yellow and blue points denote the foreground points $\mathbf{P}_{in}$ and the background points $\mathbf{P}_{out}$, respectively. At the same time, the DCS algorithm is applied to cluster the point cloud into different ring segments $\mathbf{P}^{id}$ in Fig. \ref{fig:plg2}(c). By comparing Fig. \ref{fig:plg2}(b) and Fig. \ref{fig:plg2}(c) we observe that some points belonging to the same segment are distributed inside and outside the 2D box, denoted by $\mathbf{P}_{i\_box}^{id}$ and $\mathbf{P}_{o\_box}^{id}$, respectively. We refine the labels by comparing the relative proportions of $\mathbf{P}_{i\_box}^{id}$ and $\mathbf{P}_{o\_box}^{id}$. If the number of points outside the box exceeds the number of points inside the box, we classify the $\mathbf{P}_{i\_box}^{id}$ as background points. The specific label classification is shown in the following formula:
    % plg label classification
    % {\samll
    \begin{equation}
    \label{eq:plg}
    \begin{gathered}
        \hat{y}_{id}^{ref}=\left\{
        \begin{array}{l}
	0,\,\,\,\,\,\,\text{if}\,\,prop>0.5\\
	1,\,\,\,\,\,\,\text{if}\ prop<0.1\\
	-1,\ \text{otherwise}\\
        \end{array} \right.  
                \\
        prop\ =\ \frac{\left|\mathbf{P}_{o\_box}^{id}\right|}{\left|\mathbf{P}_{i\_box}^{id}\right| + \left|\mathbf{P}_{o\_box}^{id}\right|}  
    \end{gathered}
    \end{equation}where $\hat{y}_{id}^{ref}$ represents the pseudo labels of the points belonging to the $id$-th ring segment, $\left|\cdot \right|$ denotes the number of points. We iterate over all the points to get the refined pseudo labels $\hat{\mathbf{Y}}_{ref}=\{\hat{y}_1^{ref},\dots,\hat{y}_{N_{in}}^{ref}\}\in\{-1,0,1\}^{N_{in}}$ in Fig. \ref{fig:plg2}(d), where the numbers $-1$, $0$, $1$ represent the ignored, background, and foreground labels of points, respectively.

\begin{figure}[t]
    \begin{center}
       \includegraphics[width=1.0\linewidth]{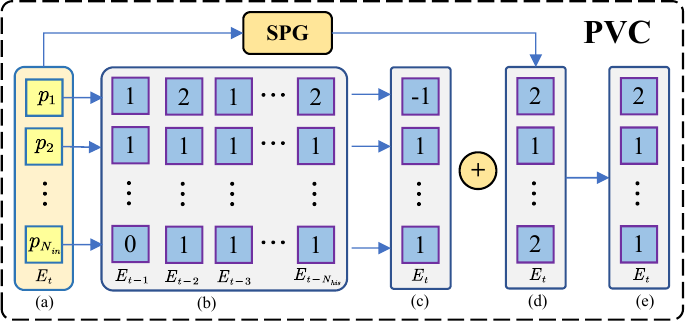}
    \end{center}
       \caption{The PVC module architecture. For the current epoch in Fig. \ref{fig:pvc}(a), historical predictions in Fig. \ref{fig:pvc}(b) are used to vote and obtain correction pseudo labels in Fig. \ref{fig:pvc}(c). Then the pseudo labels generated by SPG in Fig. \ref{fig:pvc}(d) are corrected to obtain the final pseudo labels in Fig. \ref{fig:pvc}(e).}
    \label{fig:pvc}
    \end{figure}

    After performing the aforementioned label refinement operation, we apply the CCL to all pseudo foreground points $\mathbf{P}_{fg}$ within the 2D boxes $\mathbf{B}=\{B_1, \dots, B_{N_{box}}\}$, where $N_{box}$ is the number of 2D boxes. According to Eq. \ref{eq:ccl}, we can obtain the maximum cluster $\mathbf{P}_{\{fg,b\}}'$ of the foreground points $\mathbf{P}_{\{fg,b\}}$ in the $b$-th box, and get the corresponding point pseudo labels $\hat{\mathbf{Y}}_{\{fg,b\}}$. This process generates the final semantic labels $\hat{\mathbf{Y}}_{sem}=\{\hat{y}_1^{sem},\dots,\hat{y}_{N_{in}}^{sem}\}\in\{-1,0,\dots,N_{cls}\}^{N_{in}}$ and instance labels $\hat{\mathbf{Y}}_{inst}=\{\hat{y}_1^{inst},\dots,\hat{y}_{N_{in}}^{inst}\}\in\{0,1,\dots,N_{box}\}^{N_{in}}$, where $N_{cls}$ is the number of classes, $\hat{y}_i^{sem}$ and $\hat{y}_i^{inst}$ are the $i$-th point semantic and instance labels, respectively.
    \begin{equation}
    \label{eq:ccl}
        \mathbf{P}_{\left\{fg,b\right\}}'=Max\left(CCL\left(\mathbf{P}_{\{fg,b\}}\right)\right)
    \end{equation}
    \noindent\textbf{Point-based Voting Label Correction (PVC).} To further improve the accuracy of the pseudo label generated by the SPG, we propose the PVC module. The purpose of this module is to leverage the network generalization ability to correct the pseudo labels. Specifically, we establish a matrix space $\mathbf{H}$ of dimension $N_{num} \times N_{his} \times N_{in}$ to store the teacher model predictions from the previous $N_{his}$ epochs, where $N_{num}$ is the number of the samples in the training split. As shown in Fig. \ref{fig:pvc}, for a frame of point cloud $\mathbf{P}$, when the current training epoch $E_t$ reaches the defined epoch $E_s$, we use the predictions from the previous $N_{his}$ epochs to conduct voting. In the voting process, we select the points with segmentation scores greater than $\tau_h$ or less than $\tau_l$ as reliable foreground and background, respectively, and ignore the rest. If the pseudo label of a point is deemed reliable for more than $T_h$ samples, we override its pseudo segmentation label by the voted majority.
    % pvc module
    
    \noindent\textbf{Ring Segment-based Label Correction (RSC).} The RSC module conducts voting inside every laser ring segment produced by the DCS algorithm to further refine the predicted masks. Our RSC algorithm is shown in Appendix A.1. In this algorithm, if the proportion of points in the current ring segment that belongs to the background exceeds a threshold $T_1$, we override the corresponding predicted labels as background. Similarly, if the proportion of points in the current segment for foreground exceeds a threshold $T_2$, we override the predicted labels to the current foreground class.
    % 2D Instance mask
    \begin{table*}[t]
    \centering
    \adjustbox{max width=1.0\textwidth}{
    {
    \small
    \begin{tabular}{c|c|c|c|ccc|ccc}
    \hline
    \multirow{2}{*}{Supervision} & \multirow{2}{*}{Annotation} & \multirow{2}{*}{Model}        & Epoch  & \multicolumn{3}{c|}{mAP}                         & \multicolumn{3}{c}{AP}                            \\ 
    \cline{5-10}
                                 &                             &                               & / Iter & AP             & AP50           & AP75           & Veh.           & Ped.           & Cyc.            \\ 
    \hline
    Full                         & 2D Mask                     & CondInst \shortcite{tian2020Condinst}                    & 24e    & 45.35          & 69.74          & 48.87          & 64.23          & 39.61          & 32.21           \\ 
    \hline
    \multirow{4}{*}{Weak}        & 2D Box                      & BoxInst \shortcite{Tian2020BoxInstHI}                     & 24e    & 34.61          & 65.35          & 32.45          & 48.48          & 27.65          & 27.69           \\ 
    \cline{2-10}
                                 & 2D+3D Box                   & LWSIS \shortcite{li2023lwsis}                       & 90k    & \textbf{37.77}          & -              & -              & -              & -              & -               \\ 
    \cline{2-10}
                                 & \multirow{2}{*}{2D Box}     & \multirow{2}{*}{MWSIS (ours)} & 24e    & 37.20 & 67.03 & 38.07 & \textbf{51.13} & 29.05 & 31.42  \\ 
    \cline{4-10}
                                 &                             &                               & 90k    & 37.74          & \textbf{67.35}          & \textbf{41.38}          & 49.72          & \textbf{30.18}          & \textbf{33.32}   \\
    \hline
    \end{tabular}
    % }
    }
    }
    \caption{Performance comparisons of 2D instance segmentation on Waymo val. dataset. Abbreviations: vehicle (Veh.), pedestrian (Ped.), cyclist (Cyc.). 2D Box and 3D Box representations use boxes as weakly supervised annotations.}
    \label{tab:performance 2d}
    \end{table*}
    % 3D Instance mask
    \begin{table*}[t]
    \centering
    \adjustbox{max width=1.0\textwidth}{
    {\small
    \begin{tabular}{c|c|c|ccc|ccc|cccc} 
    \hline
    \multirow{2}{*}{Supervision} & \multirow{2}{*}{Annotation} & \multirow{2}{*}{Model}                                                              & \multicolumn{3}{c|}{mAP}                         & \multicolumn{3}{c|}{AP}                          & \multicolumn{4}{c}{IoU}                                            \\ 
    \cline{4-13}
                                 &                             &                                                                                     & AP             & AP50           & AP75           & Veh.           & Ped.           & Cyc.           & mIoU           & Veh.           & Ped.           & Cyc.            \\ 
    \hline
    Full                         & 3D Mask                     & \multirow{4}{*}{\begin{tabular}[c]{@{}c@{}}SparseUNet\\ \shortcite{shi2020parta2}\end{tabular}} & 57.59          & 65.52          & 59.99          & 80.30          & 57.18          & 35.29          & 80.22          & 96.67          & 82.47          & 61.51           \\ 
    \cline{1-2}\cline{4-13}
    \multirow{4}{*}{Weak}        & 3D Box                      &                                                                                     & 48.29          & 63.49          & 55.42          & 68.95          & 45.94          & 29.99          & 72.70          & 89.68          & 73.58          & 54.85           \\ 
    \cline{2-2}\cline{4-13}
                                 & SAM                         &                                                                                     & 43.34          & 54.90          & 45.98          & \textbf{64.01}          & 39.89          & 26.13          & 75.59          & \textbf{93.47}          & 77.98          & 55.31           \\ 
    \cline{2-2}\cline{4-13}
                                 & \multirow{2}{*}{2D Box}     &                                                                                     & 34.18          & 49.95          & 38.01          & 44.18          & 36.43          & 21.94          & 63.38          & 73.69          & 71.55          & 44.91           \\ 
    \cline{3-13}
                                 &                             & MWSIS (ours)                                                                        & \textbf{46.93} & \textbf{56.58} & \textbf{49.78} & 62.46 & \textbf{45.73} & \textbf{32.59} & \textbf{75.93} & 90.45 & \textbf{79.00} & \textbf{58.33}  \\
    \hline
    \end{tabular}
    }
    }
    \caption{Performance comparisons of 3D instance and semantic segmentation on Waymo val. dataset. SAM represents the instance masks obtained using 2D GT boxes as prompts, which are adopted as weakly supervised annotations.}
    \label{tab:performance 3d}
    \end{table*}

\subsection{Consistency Sparse Cross-modal Supervision (CSCS)}
\label{CSCS}
\begin{figure}[t]
    \begin{center}
       \includegraphics[width=1.0\linewidth]{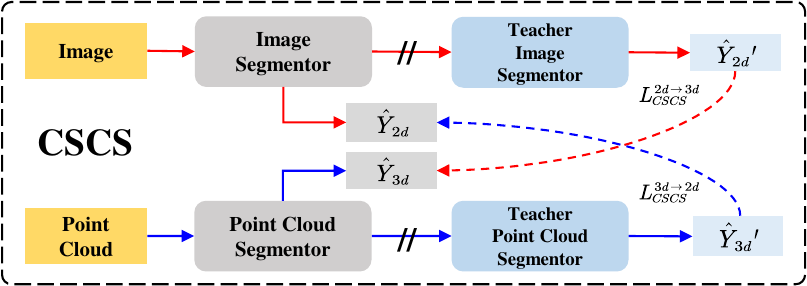}
    \end{center}
       \caption{The architecture of CSCS module.}
    \label{fig:cscs}
    \end{figure}

    Inspired by the mean teacher\cite{Tarvainen2017MeanTeacher} and CPS \cite{Chen2021CPS}, we design the CSCS module that combines the advantages of multimodal information by response distillation, as shown in Fig. \ref{fig:cscs}. For a frame of point cloud and corresponding multi-view images, we obtain two distinct point-wise predictions from the student segmentor and teacher segmentor. Then we introduce a consistency sparse loss $L_{cscs}$ to supervise the alignment between the two modalities. The CSCS loss function is defined as:
    % cscs loss
    \begin{equation}
    \label{eq:cscs}
    \begin{split}
        L_{cscs} = -\frac{1}{N_{in}N_{cls}}\sum_{i}^{N_{in}}\sum_{j}^{N_{cls}}[p^t_{i,j}\log \left(q^s_{i,j}\right)\ +\\
        \left(1-p^t_{i,j}\right)\log \left(1-q^s_{i,j}\right)]
    \end{split}
    \end{equation}
    \begin{align}
        L_{cscs}^{2d\rightarrow 3d}=L_{cscs}\left( T_{net}=2d,S_{net}=3d \right) \\
        L_{cscs}^{3d\rightarrow 2d}=L_{cscs}\left( T_{net}=3d,S_{net}=2d \right)
    \end{align}where $p_{i,j}$ and $q_{i,j}$ are the scores of the $j$-th class for the $i$-th point in different modal branches, $p^t$ and $q^s$ denote the output of teacher network ($T_{net}$) and student network ($S_{net}$), respectively. The symbols $2d$ and $3d$ represent the 2D segmentor and 3D segmentor, respectively.
\subsection{Loss}
    The overall loss function of the MWSIS is defined as:
    \begin{equation}
        L_{total}=L_{2d}+L_{3d}
    \end{equation} 
    In the 2D branch, we augment the BoxInst loss $L_{boxinst}$ with the self-supervised loss $L_{pseudo}$ and the consistency sparse loss $L_{cscs}^{3d\rightarrow2d}$. In the 3D branch, we use the classification loss $L_{cls}$, the regression loss $L_{vote}$, and the consistency sparse loss $L_{cscs}^{2d\rightarrow3d}$.
    \begin{align}
        L_{2d} &= \alpha_1 L_{boxinst} + \alpha_2 L_{pseudo} + \alpha_3 L_{cscs}^{3d\rightarrow2d} \\
        L_{3d} &= \alpha_4 L_{cls} + \alpha_5 L_{vote} + \alpha_6 L_{cscs}^{2d\rightarrow3d}
    \end{align}where $\alpha_1\sim\alpha_6$ are set as 1.0, 1.0, 0.5, 100.0, 1.0, 2.0 to balance loss terms, respectively.

\section{Experiments}
\subsection{Waymo Open Dataset}
    We conduct our experiments on version 1.4.0 of the Waymo dataset, which includes both well synchronized \& aligned LiDAR points and images. Waymo released a panoptic segmentation dataset \cite{Mei2022WaymoODPanseg}, which includes 61,480 images for training and 9,405 images for validation. For the 3D segmentation, the dataset contains 23,691 and 5,976 frames for training and validation, respectively. Due to the differences between 2D and 3D segmentation tasks on training datasets, we train and evaluate the corresponding segmentation task on two datasets, respectively. We specifically focus on the vehicle, pedestrian, and cyclist classes.
    
\subsection{Implementation Details}
    \noindent\textbf{Evaluation Metric.} For 2D and 3D instance segmentation, we follow the COCO evaluation method using AP (average precision over IoU thresholds), AP50 (IoU is 0.50), and AP75 (IoU is 0.75) to evaluate our method. For 3D semantic segmentation, we use the standard mIoU to evaluate the results.

    \noindent\textbf{Training Setting.} We train our model for 24 epochs with a batch size of 8 on 4 A6000 GPUs. Each batch contains 1 frame of the point cloud and 5 images. The 2D network adopts the SGD optimizer at a learning rate of 0.01, while the 3D network employs the AdamW optimizer with a one-cycle learning rate policy, setting the maximum rate to 0.001. Our framework is based on the mmdetection framework.

    \noindent\textbf{Data Augmentation.} In our framework, the 2D and 3D modalities are mutually independent, and we can decouple the data augmentation for multimodal data through the point-pixel mapping relationship. For the 2D branch, we employ resizing, random flipping, normalization, and padding as data augmentation methods, while for the 3D branch, we apply global rotation, global translation, global scaling, random flipping, and shuffling as data augmentation methods.
    
    \begin{table*}[t]
    \centering
    {\small
    % \adjustbox{max width=0.85\textwidth}{
    \begin{tabular}{c|c|cc|cc|cc|cc} 
    \hline
    \multirow{2}{*}{Model} & Method       & \multicolumn{2}{c|}{mAP/mAPH}           & \multicolumn{2}{c|}{\textit{Veh.} AP/APH} & \multicolumn{2}{c|}{\textit{Ped.}AP/APH} & \multicolumn{2}{c}{\textit{Cyc.}AP/APH}  \\ 
    \cline{2-10}
                           & Pre-training & L1                 & L2                 & L1                 & L2                   & L1        & L2                            & L1                 & L2                   \\ 
    \hline
    FSD                    & $\mbox{-}$   & 78.7/76.3          & 71.9/69.7          & 77.8/77.3          & 68.9/68.5            & 81.9/76.4 & 73.2/68.0                     & 76.5/75.2          & 73.8/72.5            \\ 
    \cline{2-10}
    \shortcite{fan2022fsd}               & $\checkmark$ & \textbf{79.2/76.8} & \textbf{72.4/70.2} & \textbf{78.1/77.6} & \textbf{69.3/68.9}   & 81.8/76.3 & 73.0/67.9                     & \textbf{77.6/76.5} & \textbf{74.8/73.7}   \\
    \hline
    \end{tabular}
    % }
    }
    \caption{3D detection performance on Waymo val. dataset. Our pre-training backbone improves the detection performance of the FSD.}
    \label{tab:performance detect}
    \end{table*}

    % ablation for 3D segmentation
    \begin{table*}[t]
    \centering
    % \adjustbox{max width=0.85\textwidth}{
    {\small
    \begin{tabular}{c|c|c|c|c|c|c|c|c|c} 
    \hline
    \multirow{2}{*}{Supervision} & \multirow{2}{*}{Annotation} & \multirow{2}{*}{Model} & \multicolumn{5}{c|}{Module}                                              & \multicolumn{2}{c}{Metric}       \\ 
    \cline{4-10}
                                 &                             &                        & CCL          & SPG          & PVC          & RSC          & CSCS         & mAP            & mIoU            \\ 
    \hline
    \multirow{10}{*}{Weak}       & \multirow{4}{*}{SAM}        &                        & $\mbox{-}$   & $\mbox{-}$   & $\mbox{-}$   & $\mbox{-}$   & $\mbox{-}$   & 28.57          & 70.39           \\ 
    \cline{4-10}
                                 &                             &                        & $\checkmark$ & $\mbox{-}$   & $\mbox{-}$   & $\mbox{-}$   & $\mbox{-}$   & 43.34          & 75.59           \\ 
    \cline{4-10}
                                 &                             &                        & $\checkmark$ & $\checkmark$ & $\checkmark$ & $\mbox{-}$   & $\mbox{-}$   & 45.06          & 76.63           \\ 
    \cline{4-10}
                                 &                             &                        & $\checkmark$ & $\checkmark$ & $\checkmark$ & $\checkmark$ & $\mbox{-}$   & \underline{45.54}  & \underline{77.41}   \\ 
    \cline{2-2}\cline{4-10}
                                 & \multirow{6}{*}{2D Box}     & SparseUNet             & $\mbox{-}$   & $\mbox{-}$   & $\mbox{-}$   & $\mbox{-}$   & $\mbox{-}$   & 4.63           & 42.94           \\ 
    \cline{4-10}
                                 &                             & \shortcite{shi2020parta2}                   & $\checkmark$ & $\mbox{-}$   & $\mbox{-}$   & $\mbox{-}$   & $\mbox{-}$   & 34.18          & 63.38           \\ 
    \cline{4-10}
                                 &                             &                        & $\checkmark$ & $\checkmark$ & $\mbox{-}$   & $\mbox{-}$   & $\mbox{-}$   & 42.32          & 73.87           \\ 
    \cline{4-10}
                                 &                             &                        & $\checkmark$ & $\checkmark$ & $\checkmark$ & $\mbox{-}$   & $\mbox{-}$   & 44.71          & 74.75           \\ 
    \cline{4-10}
                                 &                             &                        & $\checkmark$ & $\checkmark$ & $\checkmark$ & $\checkmark$ & $\mbox{-}$   & 45.76          & \textbf{76.29}  \\ 
    \cline{4-10}
                                 &                             &                        & $\checkmark$ & $\checkmark$ & $\checkmark$ & $\checkmark$ & $\checkmark$ & \textbf{46.93} & 75.93           \\
    \hline
    \end{tabular}
    % }
    }
    \caption{Ablation studies for 3D instance and semantic segmentation on Waymo val. dataset. }
    \label{tab:ablation for 3d}
    \end{table*}
    
\subsection{Results}
    \textbf{2D Instance Segmentation.} We compare our method with competitive fully supervised, and weakly supervised instance segmentation methods on Waymo dataset. As shown in Tab. \ref{tab:performance 2d}, our method achieves a 2.59\% mAP improvement over BoxInst and outperforms the fully supervised approach in the cyclist class. Moreover, our method achieves comparable performance to the LWSIS, which utilizes more precise guidance of points inside the 3D box.
    
    \noindent\textbf{3D Instance and Semantic Segmentation.} We compare the performance of our MWSIS with full supervision and other weak annotations. As shown in Tab. \ref{tab:performance 3d}, we use the data processed by clustering as the baseline for comparison in row 4. The MWSIS improves the baseline by 12.75\% mAP on the instance segmentation task. On the semantic segmentation task, our method achieves 94.65\% of the fully supervised performance, and surpasses the performance based on 3D box annotations in all classes.

    \noindent\textbf{SAM with 3D Segmentation.} SAM \cite{Kirillov2023SAM} uses points, boxes, masks, and texts as prompts to segment objects in the image. We use 2D boxes as our prompts to obtain masks. The mapping relationship between the point cloud and image pixels is employed to obtain masks for the point cloud. To further obtain final mask labels, we apply the CCL algorithm to the point cloud. Our model exhibits 3.59\% mAP higher than that using SAM alone.

\begin{figure}[ht]
    \begin{center}
       \includegraphics[width=1.0\linewidth]{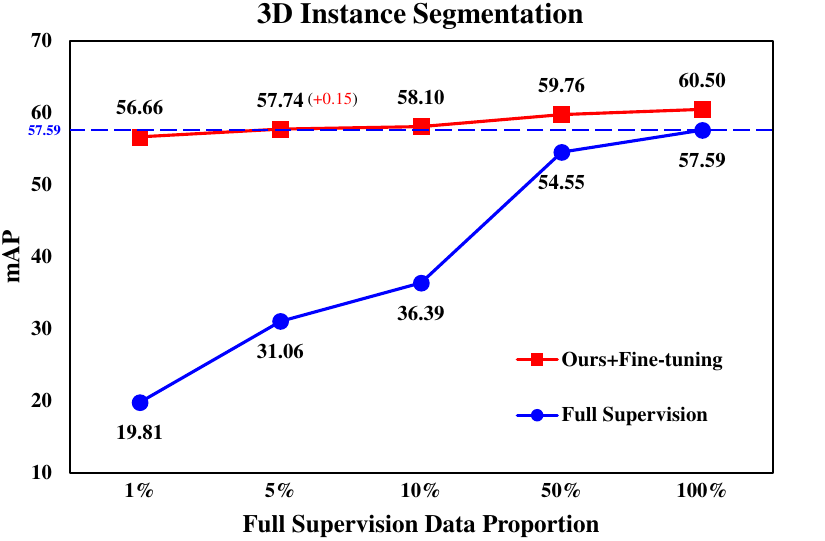}
    \end{center}
       \caption{3D instance segmentation on Waymo val. dataset. The weakly supervised pre-training backbone for fine-tuning compared to the fully supervised performance.}
    \label{fig:finetuning}
    \end{figure}

    \noindent\textbf{The Potential for Downstream Tasks.} To verify the potential of our 3D label correction method, we pre-train the 3D backbone on Waymo training set and then fine-tune it on downstream tasks. In Fig. \ref{fig:finetuning}, we compare the instance segmentation performance after fine-tuning with different proportions of full supervision data. The network outperforms full supervision with only 5\% of the full supervision data. Additionally, it is shown in Tab. \ref{tab:performance detect} that our pre-training backbone can improve the performance of 3D object detectors like FSD \cite{fan2022fsd}.
    
    \subsection{Ablation Studies}
    \textbf{3D Pseudo Label Generation.} Fig. \ref{fig:comparison of mIoU results} shows comparisons of pseudo labels generated by the SPG with other methods. It can be seen that the quality of pseudo labels generated by the SPG is higher than that of other methods, even SAM \cite{Kirillov2023SAM}. In rows 4 and 5 of Tab. \ref{tab:ablation for 3d}, the SPG performance achieves 8.14\% mAP higher than that using the CCL directly (baseline), showing that the SPG can effectively improve the quality of pseudo labels by using the spatial prior information of the point cloud. In rows 5 and 6 of Tab. \ref{tab:ablation for 3d}, the PVC has 2.32\% mAP higher than that using the SPG, which shows that the PVC corrects the noise labels in the network by voting. Compared with the 3rd and 4th rows, 8th and 9th rows of Tab. \ref{tab:ablation for 3d}, the performance improves by 0.48\% mAP and 1.05\% mAP, respectively, after incorporating the RSC module. These experiments reveal that the depth prior ring segment can effectively correct the predictions.

    \noindent\textbf{Consistency Sparse Cross-modal Supervision (CSCS).} In Tab. \ref{tab:cscs}, after applying the CSCS, the segmentation performances of 2D and 3D networks are improved by 0.43\% mAP and 1.17\% mAP, respectively, which confirms the effectiveness of multimodal response distillation.
    % CSCS
    \begin{table}[ht]
    \centering
    % \adjustbox{max width=1.0\columnwidth}{
    {\small
    \begin{tabular}{c|c|c|c|c} 
    \hline
    Model                                                                       & Task                & $L_{cscs}^{2d\rightarrow 3d}$ & $L_{cscs}^{3d\rightarrow 2d}$ & mAP             \\ 
    \hline
    BoxInst                                                                     & \multirow{3}{*}{2D} & $\mbox{-}$                & $\mbox{-}$                & 36.77           \\ 
    \cline{3-5}
    \shortcite{Tian2020BoxInstHI}                                                                      &                     & $\mbox{-}$                & $\checkmark$              & 36.99           \\ 
    \cline{3-5}
    + IPG                                                                       &                     & $\checkmark$              & $\checkmark$              & \textbf{37.20}  \\ 
    \hline
    \multirow{3}{*}{\begin{tabular}[c]{@{}c@{}}SparseUNet\\ \shortcite{shi2020parta2}\end{tabular}} & \multirow{3}{*}{3D} & $\mbox{-}$                & $\mbox{-}$                & 45.76           \\ 
    \cline{3-5}
                                                                                &                     & $\checkmark$              & $\mbox{-}$                & 46.37           \\ 
    \cline{3-5}
                                                                                &                     & $\checkmark$              & $\checkmark$              & \textbf{46.93}  \\
    \hline
    \end{tabular}
        % }
    }
    \caption{Comparison of different supervision methods.}
    \label{tab:cscs}
    \end{table}
    
\section{Conclusion}
    This paper proposes a novel multimodal instance segmentation framework with weak annotations. It explores two questions: one is how to correct pseudo labels with solely 2D box annotations, and the other is how to train multimodal segmentors simultaneously. To address the former challenge, we introduce various fine-grained pseudo label generation and correction methods to improve the quality of pseudo labels. For the second problem, our framework decouples the two modal branches and trains the multimodal segmentors simultaneously. Additionally, a cross-modal supervision module CSCS is proposed to utilize the teacher segmentors' output for response distillation by the point-pixel mapping relationship. By these methods, the multimodal segmentors achieve efficient training and reliable performance. Compared to the baseline, our method achieves 2.59\% mAP and 12.75\% mAP improvement on the 2D and 3D instance segmentation tasks, respectively. Moreover, the 3D segmentor outperformes the full supervision using only 5\% of the fully supervised annotations, and can also serve as a pre-training method to improve the performance on the other 3D downstream tasks.
%------------------------------------------------------------------------
% \clearpage
\section*{Acknowledgments}
    This work was supported in part by the Youth Innovation Promotion Association CAS (CX2100060053), and the Natural Science Foundation of Anhui Province under Grant 2208085J17. We sincerely appreciate all the reviewers. They give positive and high-quality comments on our paper with a lot of constructive feedback to help us improve the paper.
%------------------------------------------------------------------------
\bibliography{aaai24}
% \bibliography{Formatting-Instructions-LaTeX-2024.bbl}
%------------------------------------------------------------------------
\clearpage
% \section*{Appendix}
\appendix
%------------------------------------------------------------------------
\section{Appendix}
Our supplemental file contains the following materials:
    \begin{itemize}
        \item \textbf{Algorithm Implementation.} In Section \ref{alg:alg}, we provide algorithmic implementations of the DCS algorithm and the RSC algorithm.
        \item \textbf{Experiment Setting.} In Section \ref{Experiment Setting}, we provide the detailed setup for training, module parameters.
        \item \textbf{Detailed Results.} In Section \ref{Detailed Results}, we provide the detailed experiment results.
        \item \textbf{Detailed Ablation Experiments.} In Section \ref{detailed ablation}, we supplement detailed ablation experiments on the IPG, the SPG, the PVC, the RSC, and other methods.
    \end{itemize}
\subsection{Algorithm}
\label{alg:alg}
\label{apd:DCS}
    \textbf{Depth Clustering Segmentation (DCS).} In a single scan, a LiDAR sensor captures a frame of point cloud data in the form of an $M \times N$ measurement matrix known as a range image, where $M$ represents the number of beams and $N$ represents the number of measurements. The DCS algorithm partitions the point cloud into ring segments based on the smooth variations in depth, resulting in a more organized and structured representation of the point cloud.

\begin{figure*}[t]
\begin{center}
   \includegraphics[width=0.8\linewidth]{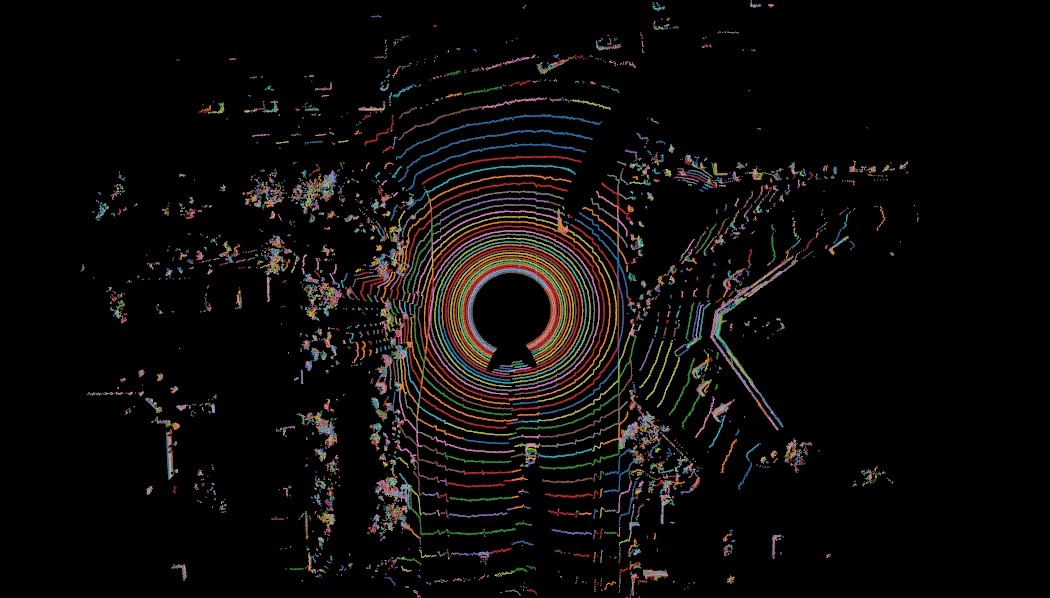}
\end{center}
   \caption{The ring segments for the point cloud. Different colors represent different segments.}
\label{fig:ring segment}
\end{figure*}

    To achieve this, the DCS algorithm traverses each row of the point cloud and records the depth of each column. When the depth changes smoothly, the current point belongs to the same ring segment. Otherwise, it belongs to a different segment. Fig. \ref{fig:ring segment} shows the results of applying the DCS algorithm to a frame point cloud.
    
    In the point cloud scene, the density of points decreases as the distance increases. Therefore, we set the window size and the depth threshold to dynamic thresholds based on the distance of the point cloud. Algorithm \ref{alg:simplified dcs} provides a concise overview of the DCS, while algorithm \ref{alg:detailed dcs} presents a detailed implementation of the DCS.

    % Simplified dcs algorithm
    \begin{algorithm}[H]
    \small
    \footnotesize
    \SetAlgoLined
    \SetKwInOut{Input}{Input}
    \SetKwInOut{Output}{Output}
        \Input{Range image $\mathbf{D}\in\mathbb{R}^{M\times N}$. \\
            Coordinates $\mathbf{C}\in \mathbb{R}^{N_{in}\times 2}$ of points on a range image. \\
            Cluster id matrix $\mathbf{ID}=\{0\}^{M\times N}$. \\
            Depth threshold $T$. \\
            Number of clusters $N_{clu}=0$. \\}
        \Output{Ring segments $\mathbf{R}$.}
        \For{$r=1:M$}
            {
                \For{$i=1:N$}
                {
                    \If{$\left|\mathbf{D}\left[r\right]\left[i\right]-\mathbf{D}\left[r\right]\left[i-1\right]\right|<T$ {\bf and} {$i\ge 2$}}
                    {                        $\mathbf{ID}\left[r\right]\left[i\right]=\mathbf{ID}\left[r\right]\left[i-1\right]$;
                    }
                    \Else
                    {{$\mathbf{ID}\left[r\right]\left[i\right]=N_{clu}$;\\}
                    {$N_{clu} += 1$;}}
                }
            }
        % \tcp{Use point coordinates in the $\mathbf{D}$ to obtain the corresponding cluster id}
        $\mathbf{R}=\mathbf{ID}[\mathbf{C}]$
    \caption{Simplified DCS}
    \label{alg:simplified dcs}
    \end{algorithm}
    % Detailed dcs algorithm
    \begin{algorithm}[H]
        \small
        \SetAlgoLined
        \SetKwInOut{Input}{Input}
        \SetKwInOut{Output}{Output}
        \Input{Range image $\mathbf{D}\in\mathbb{R}^{M\times N}$. \\
            Coordinates $\mathbf{C}\in \mathbb{R}^{N_{in}\times 2}$ of points on a range image. \\
            Equal table $\mathbf{E}=$\\ $[[0,1,\dots,N-1],\dots,[0,1,\dots,N-1]]\in \mathbb{R}^{M\times N}$. \\
            % Equal table $\mathbf{E}=\left[[0,1,\dots,N-1],\dots,[0,1,\dots,N-1]\right]\in \mathbb{R}^{M\times N}$. \\
            % Cluster id matrix $\mathbf{ID}=\{0\}^{M\times N}$. \\
            % Depth threshold $T$. \\
            % Window size $W$. \\
            % Number of clusters $N_{clu}=0$.
            }
        \Output{Ring segments $\mathbf{R}\in \mathbb{R}^{N_{in}}$.}
        \For{$r=1:M$}
            {
                {$M_r = max(\mathbf{D}[r,:])$}
                \tcp*{Get the maximum depth of points on the $r$-th beam.}
                {$W_r = \frac{50}{M_r}\times W$}
                \tcp*{Get the window size.}
                {$T_r = \frac{M_r}{50}\times 50$}
                \tcp*{Get the depth threshold.}
                \tcp{Build equal table.}
                \For{$i=1:N$}
                {
                    \If{$\mathbf{D}[r][i] \ne Nan$}
                    {
                        \For{$j=1:\frac{W_r}{2}$}
                        {
                            \If{$i\ge j$ {\bf and} $\mathbf{D}[r][i-j]\ne Nan$ {\bf and} {$|\mathbf{D}[r][i-j]-\mathbf{D}[r][i]|<T_r$}}
                            {
                                $\mathbf{E}[i]=\mathbf{E}[i-j]$;\\
                                \textbf{break}
                            }
                            
                        }
                    }
                }
                \tcp{Assign cluster id.}
                \For{$i=1:N$}
                {
                    \If{$\mathbf{E}[r][i] \ne Nan$}
                    {
                        \If{$\mathbf{E}[r][i]=i$}
                        {
                            {$\mathbf{ID}[r][i]=N_{clu}$;}\\
                            {$N_{clu} += 1$;}\\
                        }
                        
                    }
                }
                \tcp{Relabel.}
                \For{$i=1:N$}
                {
                    \If{$\mathbf{D}[r][i]\ne Nan$}
                    {
                        {$label=i$;}\\
                        \While{$label\ne \mathbf{E}[r][label]$}
                        {
                            $lable=\mathbf{E}[r][label]$;
                        }
                        $\mathbf{ID}[r][i]=\mathbf{ID}[r][label]$;
                    }
                }
        }
        \tcp{Use point coordinates in the $\mathbf{D}$ to obtain the corresponding cluster ids.}
        {$\mathbf{R}=\mathbf{ID}[\mathbf{C}]$}\\
    \caption{DCS}
    \label{alg:detailed dcs}
    \end{algorithm}
    
    \noindent\textbf{Ring Segment-based Label Correction (RSC).} The ring segment serves as a novel representation of the point cloud, where different instances are composed of distinct ring segments. If the same ring segment belongs to the different instances or semantics, the predictions can be corrected by voting, thereby improving the accuracy of the predicted labels. Algorithm \ref{alg:rsc} provides the overview of the RSC, where $\left|\cdot \right|$ denotes the number of points.
    \label{apd:rsc}
    %  RSC alg
    \begin{algorithm}[H]
        \small
        \SetAlgoLined
        \SetKwInOut{Input}{Input}
        \SetKwInOut{Output}{Output}
        \Input{Predicted labels $\hat{\mathbf{Y}}_{pred}\in\mathbb{R}^{N_{in}}$.\\
            Point ring segments $\mathbf{R}\in\mathbb{R}^{N_{in}}$.\\
            Class labels $CLS\in\mathbb{R}^{N_{cls}}.$\\
            Thresholds $T_1,T_2$.}
        \Output{$\hat{\mathbf{Y}}_{pred,out}.$}    
        {$\hat{\mathbf{Y}}_{pred,out} = \hat{\mathbf{Y}}_{pred};$}\\
        {$Mask_{bg}=\hat{\mathbf{Y}}_{pred}==0;$}\\
        \For{$i=1:N_{cls}$}
        {
            {$Mask_i=\hat{\mathbf{Y}}_{pred}==CLS_i;$}\\
            \tcp*[f]{Remove duplicate ring segments.}\\
            {$R_i=Set\left(R\left[Mask_i\right]\right);$}\\
            \For{$j=1:\left|R_i\right|$}
            {
                \If{$\left|R\left[Mask_{bg}\right]==R_{i,j}\right|/\left|R[Mask_{i}]==R_{i,j}\right|>T_1$}{
                    $\hat{\mathbf{Y}}_{pred,out}\left[R==R_{i,j}\right]=0;$\\
                }
                \ElseIf{$\left|R[Mask_{i}]==R_{i,j}\right|/\left|R==R_{i,j}\right|>T_2$}{
                    $\hat{\mathbf{Y}}_{pred,out}\left[R==R_{i,j}\right]=CLS_i;$
                }
            }
        }
        \caption{RSC}
        \label{alg:rsc}
        \end{algorithm}

\subsection{Experiment Setting}
\label{Experiment Setting}
    \textbf{Training Setup.} In the 2D branch, we do not modify the hyperparameter configuration of BoxInst, such as the convolution kernel size and stride. In the 3D branch, we employ the SparseUNet as our 3D backbone, consisting of 5 encoder and 5 decoder layers. Additionally, a segmentation head is utilized to predict the semantic label and the center of points for subsequent clustering.

    % We train our model for 24 epochs with a batch size of 8 on 4 A6000 GPUs. Each batch contains 1 frame of the point cloud and 5 images. The 2D network adopts the SGD optimizer at a learning rate of 0.01, while the 3D network employs the AdamW optimizer with a one-cycle learning rate policy, setting the maximum rate to 0.001.
    
    \noindent\textbf{Module Setup.} In the IPG module, binarization thresholds are set to 0.3 and 0.7, while the correlation coefficient $k$ is set to 1. In the SPG module, the distance thresholds are set as 0.6 m, 0.1 m, and 0.15 m for the vehicle, pedestrian, and cyclist, respectively. In the DCS algorithm, the depth threshold is set to 0.24 m with a window size of 10 at 50 meters. In the PVC module, the number of historical votes $N_{his}$ is set to 4, and the confidence threshold of reliable points is set to 0.5. In the RSC algorithm, $T_1$ and $T_2$ are set as 0.5 and 0.7, respectively.
    
    % \noindent\textbf{Data Augmentation.} In our framework, the 2D and 3D modalities are mutually independent, and we can decouple the data augmentation for multimodal data through the point-pixel mapping relationship. For the 2D branch, we employ resizing, random flipping, normalization, and padding as data augmentation methods, while for the 3D branch, we apply global rotation, global translation, global scaling, random flipping, and shuffling as data augmentation methods.
    % 2D Instance mask
    \begin{table*}[t]
    \centering
    \adjustbox{max width=1.0\textwidth}{
    \begin{tabular}{c|c|c|c|ccc|ccc|ccc|ccc} 
    \hline
    \multirow{2}{*}{Supervision} & \multirow{2}{*}{Annotation} & \multirow{2}{*}{Model} & Epoch   & \multicolumn{3}{c|}{mAP}                         & \multicolumn{3}{c|}{Veh.}                     & \multicolumn{3}{c|}{Ped.}                  & \multicolumn{3}{c}{Cyc.}                       \\ 
    \cline{5-16}
                                 &                             &                        & / Iter & AP             & AP50           & AP75           & AP             & AP50           & AP75           & AP             & AP50           & AP75           & AP             & AP50           & AP75            \\ 
    \hline
    Full                   & 2D Mask                     & CondInst \shortcite{tian2020Condinst}               & 24e     & 45.35          & 69.74          & 48.87          & 64.23          & 84.81          & 70.69          & 39.61          & 74.24          & 37.46          & 32.21          & 50.17          & 38.47           \\ 
    \hline
    \multirow{4}{*}{Weak} & 2D Box                      & BoxInst \shortcite{Tian2020BoxInstHI}               & 24e     & 34.61          & 65.35          & 32.45          & 48.48          & 79.18          & 53.76          & 27.65          & 66.81          & 16.55          & 27.69          & 50.05          & 27.03           \\ 
    \cline{2-16}
                                 & 2D+3D Box                   & LWSIS \shortcite{li2023lwsis}                  & 90k     & 37.77          & -              & -              & -              & -              & -              & -              & -              & -              & -              & -              & -               \\
    \cline{2-16}
                                 & \multirow{2}{*}{2D Box}     & \multirow{2}{*}{MWSIS (ours)} & 24e     & 37.20 & 67.03 & 38.07 & 51.13 & 81.53 & 57.57 & 29.05 & 68.86 & 17.78 & 31.42 & 50.71 & 38.87  \\ 
    \cline{4-16}
                                 &                             &                        & 90k     & 37.74          & 67.35          & 41.38          & 49.72          & 80.66          & 55.52          & 30.18          & 68.13          & 22.19          & 33.32  & 53.27  & 46.43   \\
    \hline
    \end{tabular}
    }
    \caption{Detailed performance comparisons of 2D instance segmentation on Waymo val. dataset.}
    \label{tab:detailed performance 2d}
    \end{table*}
    % 3D Instance mask
    \begin{table*}[t]
    \centering
    \adjustbox{max width=1.0\textwidth}{
    \begin{tabular}{c|c|c|ccc|ccc|ccc|ccc|cccc} 
    \hline
    \multirow{2}{*}{Supervision} & \multirow{2}{*}{Annotation} & \multirow{2}{*}{Model}                                                    & \multicolumn{3}{c|}{mAP}                         & \multicolumn{3}{c|}{Veh.}                     & \multicolumn{3}{c|}{Ped.}                  & \multicolumn{3}{c|}{Cyc.}                     & \multicolumn{4}{c}{IoU}                                            \\ 
    \cline{4-19}
                                 &                             &                                                                           & AP             & AP50           & AP75           & AP             & AP50           & AP75           & AP             & AP50           & AP75           & AP             & AP50           & AP75           & mIoU           & Veh.        & Ped.     & Cyc.         \\ 
    \hline
    Full                   & 3D Mask                     & \multirow{4}{*}{\begin{tabular}[c]{@{}c@{}}SparseUNet\\ \shortcite{shi2020parta2}\end{tabular}} & 57.59          & 65.52          & 59.99          & 80.30          & 87.23          & 82.71          & 57.18          & 68.47          & 59.87          & 35.29          & 40.86          & 37.37          & 80.22          & 96.67          & 82.47          & 61.51           \\ 
    \cline{1-2}\cline{4-19}
    \multirow{4}{*}{Weak} & 3D Box                      &                                                                           & 48.29          & 63.49          & 55.42          & 68.95          & 85.83          & 78.96          & 45.94          & 63.49          & 51.76          & 29.99          & 41.15          & 35.54          & 72.70          & 89.68          & 73.58          & 54.85           \\ 
    \cline{2-2}\cline{4-19}
                                 & SAM                  &                                                                           & 43.34          & 54.90          & 45.98          & 64.01          & 76.23          & 66.87          & 39.89          & 54.00          & 42.49          & 26.13          & 34.48          & 28.58          & 75.59          & 93.47          & 77.98          & 55.31           \\ 
    \cline{2-2}\cline{4-19}
                                 & \multirow{2}{*}{2D Box}     &                                                                           & 34.18          & 49.95          & 38.01          & 44.18          & 66.73          & 49.90          & 36.43          & 49.10          & 39.71          & 21.94          & 34.03          & 24.42          & 63.38          & 73.69          & 71.55          & 44.91           \\ 
    \cline{3-19}
                                 &                             & MWSIS (ours)                                                                     & 46.93 & 56.58 & 49.78 & 62.46 & 76.78 & 67.12 & 45.73 & 55.98 & 48.30 & 32.59 & 36.96 & 33.93 & 75.93 & 90.45 & 79.00 & 58.33  \\
    \hline
    \end{tabular}
    }
    \caption{Detailed performance comparisons of 3D instance and semantic segmentation on Waymo val. dataset.}
    \label{tab:detailed performance 3d}
    \end{table*}
\subsection{Detailed Results.}
\label{Detailed Results}
    In Tab. \ref{tab:detailed performance 2d} and Tab. \ref{tab:detailed performance 3d} we provide detailed evaluation results on Waymo validation set.
    
    As shown in Tab. \ref{tab:detailed performance 2d}, our method achieves a 2.59\% mAP improvement over the BoxInst, and outperforms the fully supervised approach in the cyclist class. Moreover, our method achieves comparable performance to the LWSIS, which utilizes more precise guidance of points inside the 3D box.

    As shown in Tab. \ref{tab:detailed performance 3d}, we use the data processed by clustering as the baseline for comparison in row 4. The MWSIS improves the baseline by 12.75\% mAP on the 3D instance segmentation task. On the 3D semantic segmentation task, our method achieves 94.65\% of the fully supervised performance, and surpasses the performance based on 3D box annotations in all classes.

\subsection{Ablation Experiments}
\label{detailed ablation}
    \textbf{Instance-based Pseudo Mask Generation (IPG).} Tab. \ref{tab:ipg ablation} compares the impacts of the mask score, IoU, and centrality weight hyperparameter $k$ on the 2D instance segmentation task. If we only focus on the centrality relationship between the predicted boxes and GT boxes, the performance improves by 1.11\% mAP. If we further consider the reliability of predicted masks, the performance increases by 2.16\% mAP. These experiments reveal that higher confidence and higher match predictions are more effective for the self-supervision.
    % IPG
    \begin{table}[t]
    \centering
    \adjustbox{max width=0.8\columnwidth}{
    \begin{tabular}{c|c|c|c|cccc} 
    \hline
    \multirow{2}{*}{Model}                                                       & \multicolumn{3}{c|}{IPG}                 & \multicolumn{4}{c}{AP}                                             \\ 
    \cline{2-8}
                                                                                 & Score        & IoU          & $k$          & mAP            & Veh.        & Ped.     & Cyc.         \\ 
    \hline
    \multirow{5}{*}{\begin{tabular}[c]{@{}c@{}}BoxInst\\\shortcite{Tian2020BoxInstHI}\end{tabular}} & $\mbox{-}$   & $\mbox{-}$   & $\mbox{-}$ & 34.61          & 48.48          & 27.65          & 27.69           \\ 
    \cline{2-8}
                                                                                 & $\mbox{-}$   & $\checkmark$ & 1          & 35.72          & 50.47          & 28.85          & 27.83           \\ 
    \cline{2-8}
                                                                                 & $\checkmark$ & $\checkmark$ & $\mbox{-}$ & 36.49          & 50.53          & 28.61          & 30.33           \\ 
    \cline{2-8}
                                                                                 & $\checkmark$ & $\checkmark$ & 1          & \textbf{36.77} & 50.36          & 28.41          & \textbf{31.53}  \\ 
    \cline{2-8}
                                                                                 & $\checkmark$ & $\checkmark$ & 5          & 35.62          & \textbf{50.58} & \textbf{28.87} & 27.42           \\
    \hline
    \end{tabular}
    }
    \caption{Ablation studies for the IPG model.}
    \label{tab:ipg ablation}
    \end{table}
    % boxinst ablation
    \begin{table}[t]
    \centering
    \adjustbox{max width=0.8\columnwidth}{
    \begin{tabular}{c|c|c|cccc} 
    \hline
    Model                    & IPG & CSCS & mAP   & Veh.   & Ped.   & Cyc.    \\ 
    \hline
    \multirow{4}{*}{\begin{tabular}[c]{@{}c@{}}BoxInst\\\shortcite{Tian2020BoxInstHI}\end{tabular}} & $\mbox{-}$   & $\mbox{-}$    & 34.61 & 48.48 & 27.65 & 27.69  \\ 
    \cline{2-7}
                             & $\checkmark$   & $\mbox{-}$    & 36.77 & 50.36 & 28.41 & \textbf{31.53}  \\ 
    \cline{2-7}
                             & $\mbox{-}$   & $\checkmark$    & 36.65 & \textbf{51.36} & \textbf{31.83} & 26.75  \\ 
    \cline{2-7}
                             & $\checkmark$   & $\checkmark$    & \textbf{37.20} & 51.13 & 29.05 & 31.42  \\
    \hline
    \end{tabular}
        }
    \caption{Comparisons of different methods for BoxInst.}
    \label{tab:boxinst ablation}
    \end{table}
  
    \noindent\textbf{BoxInst Performance with Different Modules.} In Tab. \ref{tab:boxinst ablation}, we can observe that the use of the IPG leads to a significant improvement in the few-shot cyclist class, with an increase of 3.84\% mAP. The exploitation of the CSCS results in notable improvements in the vehicle and pedestrian classes, with increases of 2.88\% mAP and 4.18\% mAP, respectively. By combining both of these modules, we achieve better overall performance.
    
    % pvc ablation
    \begin{table*}[t]
    \centering
    \small
    \adjustbox{max width=1.0\textwidth}{
    \begin{tabular}{c|c|cccc|cccc} 
    \hline
    \multirow{2}{*}{Model}                                                  & PVC  & \multicolumn{4}{c|}{AP}                                  & \multicolumn{4}{c}{IoU}                                            \\ 
    \cline{2-10}
                                                                            & vote & mAP            & Veh.            & Ped.   & Cyc.            & mIoU           & Veh.            & Ped.            & Cyc.             \\ 
    \hline
    \multirow{4}{*}{\begin{tabular}[c]{@{}c@{}}SparseUNet\\\shortcite{shi2020parta2}\end{tabular}} & 0    & 42.32          & 55.77          & 40.36 & 30.83          & 73.87          & 87.35          & 78.01          & 56.25           \\ 
    \cline{2-10}
                                                                            & 3    & 44.44          & 57.52          & 44.57 & 31.24          & 73.46          & 85.92          & 77.63          & 56.82           \\ 
    \cline{2-10}
                                                                            & 4    & \textbf{44.71} & 57.43          & 44.08 & \textbf{32.63} & 74.75          & 87.33          & 78.64          & \textbf{58.29}  \\ 
    \cline{2-10}
                                                                            & 5    & 44.39          & \textbf{57.92} & 43.69 & 31.57          & \textbf{75.28} & \textbf{88.47} & \textbf{79.38} & 57.99           \\
    \hline
    \end{tabular}
        }
    \caption{Performance comparisons of the PVC on Waymo val. dataset. The vote represents the number of historical predictions.}
    \label{tab:pvc ablation}
    \end{table*}
    
    \noindent\textbf{Point-based Voting Label Correction (PVC).} We conduct ablation studies to examine the impacts of different numbers of historical votes on the 3D instance segmentation task. As shown in Tab. \ref{tab:pvc ablation}, we achieve 2.12\% mAP and 0.88\% mIoU performance improvement, when utilizing 4 votes.
    
    % RSC ablation
    \begin{table*}[t]
    \centering
    \small
    \adjustbox{max width=0.68\textwidth}{
    \begin{tabular}{c|c|c|cccc|cccc} 
    \hline
    \multirow{2}{*}{Supervison}                                           & \multirow{2}{*}{Annotation}                                            & \multirow{2}{*}{RSC} & \multicolumn{4}{c|}{AP}                                           & \multicolumn{4}{c}{IoU}                                            \\ 
    \cline{4-11}
                                                                          &                                                                        &                      & mAP            & Veh.            & Ped.            & Cyc.            & mIoU           & Veh.            & Ped.            & Cyc.             \\ 
    \hline
    \multirow{2}{*}{Full} & \multirow{2}{*}{3D Mask}                                                  & $\mbox{-}$           & 57.59          & 80.30          & 57.18          & 35.29          & 80.22          & 96.67          & 82.47          & 61.51           \\ 
    \cline{3-11}
                                                                          &                                                                        & $\checkmark$         & \textbf{57.78} & \textbf{80.48} & 57.18          & \textbf{35.68} & \textbf{80.38} & \textbf{96.76} & 82.47          & \textbf{61.92}  \\ 
    \hline
    \multirow{6}{*}{Weak} & \multirow{2}{*}{3D Box}                                                & $\mbox{-}$           & 48.29          & 68.95          & 45.94          & 29.99          & 72.61          & 89.68          & 73.58          & 54.58           \\ 
    \cline{3-11}
                                                                          &                                                                        & $\checkmark$         & \textbf{52.28} & \textbf{74.79} & \textbf{46.20} & \textbf{35.85} & \textbf{75.53} & \textbf{93.52} & \textbf{73.84} & \textbf{59.24}  \\ 
    \cline{2-11}
                                                                          & \multirow{2}{*}{SAM} & $\mbox{-}$           & 43.34          & 64.01          & 39.89          & 26.13          & 75.59          & 93.47          & 77.98          & 55.31           \\ 
    \cline{3-11}
                                                                          &                                                                        & $\checkmark$         & \textbf{43.72} & \textbf{64.42} & 39.81          & \textbf{26.94} & \textbf{76.47} & \textbf{93.98} & 77.96          & \textbf{57.47}  \\ 
    \cline{2-11}
                                                                          & \multirow{2}{*}{2D Box}                                                & $\mbox{-}$           & 44.71          & 57.43          & 44.08          & 32.63          & 74.75          & 87.33          & 78.64          & 58.29           \\ 
    \cline{3-11}
                                                                          &                                                                        & $\checkmark$         & \textbf{45.76} & \textbf{59.95} & \textbf{44.68} & \textbf{32.66} & \textbf{76.30} & \textbf{89.50} & \textbf{78.94} & \textbf{60.44}  \\
    \hline
    \end{tabular}
    }
    \caption{Performance comparisons of the RSC on Waymo val. dataset.}

    \label{tab:rsc ablation}
    \end{table*}
   
    \noindent\textbf{Ring Segment-based Label Correction (RSC).} In Tab. \ref{tab:rsc ablation}, the RSC not only enhances the performance of weakly supervised methods, but also affects fully supervised methods. Specifically, for 2D box annotations, our method leads to improvements of 1.05\% mAP and 1.55\% mIoU. For pseudo masks generated by the SAM, the performance improves by 0.34\% mAP and 0.88\% mIoU. When using costly 3D box annotations, the performance increases by 3.99\% mAP and 2.91\% mIoU. Moreover, for the fully supervised method, we achieve an increase of 0.19\% mAP and 0.16\% mIoU. The above experiments reveal that the RSC algorithm can effectively correct the predicted labels of points.
    
    \noindent\textbf{SAM with 3D Label Correction.} In Tab. \ref{tab:detailed ablation for 3d} we supplement detailed experiments for the SAM with our methods. Compared with the methods that directly using the SAM to process the point cloud data, our method achieves 2.2\% mAP and 1.82\% mIoU performance improvements on the instance segmentation and semantic segmentation tasks, respectively.
    % ablation for 3D segmentation
    \begin{table*}[t]
    \centering
    \adjustbox{max width=1.0\textwidth}{
    \begin{tabular}{c|c|c|c|c|c|c|c|cccc|cccc} 
    \hline
    \multirow{2}{*}{Supervision}                                           & \multirow{2}{*}{Annotation}                                            & \multirow{2}{*}{Model}                                                   & \multicolumn{5}{c|}{Module}                                              & \multicolumn{4}{c|}{AP}                                           & \multicolumn{4}{c}{IoU}                                            \\ 
    \cline{4-16}
                                                                           &                                                                        &                                                                          & CCL          & SPG          & PVC          & RSC          & CSCS         & mAP            & Veh.            & Ped.            & Cyc.            & mIoU           & Veh.            & Ped.            & Cyc.             \\ 
    \hline
    \multirow{10}{*}{Weak} & \multirow{4}{*}{SAM} & \multirow{10}{*}{\begin{tabular}[c]{@{}c@{}}SparseUNet\\ \shortcite{shi2020parta2}\end{tabular}} & $\mbox{-}$   & $\mbox{-}$   & $\mbox{-}$   & $\mbox{-}$   & $\mbox{-}$   & 28.57          & 47.03          & 14.33          & 24.35          & 70.39          & 85.10          & 75.38          & 50.70           \\ 
    \cline{4-16}
                                                                           &                                                                        &                                                                          & $\checkmark$ & $\mbox{-}$   & $\mbox{-}$   & $\mbox{-}$   & $\mbox{-}$   & 43.34          & 64.01          & 39.89          & 26.13          & 75.59          & 93.47          & 77.98          & 55.31           \\ 
    \cline{4-16}
                                                                           &                                                                        &                                                                          & $\checkmark$ & $\checkmark$ & $\checkmark$ & $\mbox{-}$   & $\mbox{-}$   & 45.06          & 66.61          & \underline{42.54}  & 26.03          & 76.63          & 94.46          & 80.21          & 55.23           \\ 
    \cline{4-16}
                                                                           &                                                                        &                                                                          & $\checkmark$ & $\checkmark$ & $\checkmark$ & $\checkmark$ & $\mbox{-}$   & \underline{45.54}  & \underline{67.04}  & 42.46          & \underline{27.12}  & \underline{77.41}  & \underline{94.81}  & \underline{80.22}  & \underline{57.20}   \\ 
    \cline{2-2}\cline{4-16}
                                                                           & \multirow{6}{*}{2D Box}                                                &                                                                          & $\mbox{-}$   & $\mbox{-}$   & $\mbox{-}$   & $\mbox{-}$   & $\mbox{-}$   & 4.63           & 6.03           & 0.33           & 7.51           & 42.94          & 55.70          & 40.49          & 32.63           \\ 
    \cline{4-16}
                                                                           &                                                                        &                                                                          & $\checkmark$ & $\mbox{-}$   & $\mbox{-}$   & $\mbox{-}$   & $\mbox{-}$   & 34.18          & 44.18          & 36.43          & 21.94          & 63.38          & 73.69          & 71.55          & 44.91           \\ 
    \cline{4-16}
                                                                           &                                                                        &                                                                          & $\checkmark$          & $\checkmark$          & $\mbox{-}$          & $\mbox{-}$          & $\mbox{-}$          & 42.32          & 55.77          & 40.36          & 30.83          & 73.87          & 87.35          & 78.01          & 56.25           \\ 
    \cline{4-16}
                                                                           &                                                                        &                                                                          & $\checkmark$ & $\checkmark$ & $\checkmark$ & $\mbox{-}$   & $\mbox{-}$   & 44.71          & 57.43          & 44.08          & 32.63          & 74.75          & 87.33          & 78.64          & 58.29           \\ 
    \cline{4-16}
                                                                           &                                                                        &                                                                          & $\checkmark$ & $\checkmark$ & $\checkmark$ & $\checkmark$ & $\mbox{-}$   & 45.76          & 59.95          & 44.68          & \textbf{32.66} & \textbf{76.29} & 89.50          & 78.94          & \textbf{60.44}  \\ 
    \cline{4-16}
                                                                           &                                                                        &                                                                          & $\checkmark$ & $\checkmark$ & $\checkmark$ & $\checkmark$ & $\checkmark$ & \textbf{46.93} & \textbf{62.46} & \textbf{45.73} & 32.59          & 75.93          & \textbf{90.45} & \textbf{79.00} & 58.33           \\
    \hline
    \end{tabular}
    }
    \caption{Detailed ablation studies for 3D instance and semantic segmentation on Waymo val. dataset.}
    \label{tab:detailed ablation for 3d}
    \end{table*}
    
\begin{figure}[t]
    \centering
       \includegraphics[width=1.0\linewidth]{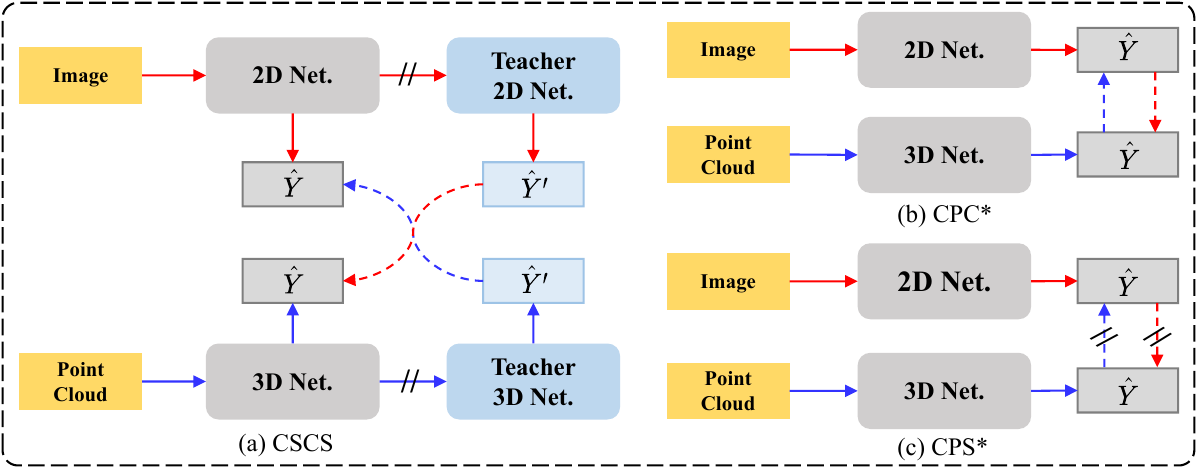}
       \caption{Different cross-supervision methods.}
    \label{fig:cscs2}
    \end{figure}

    \noindent\textbf{Comparisons of Different Cross-Supervision Methods.} Tab. \ref{tab:cscs1} compares the performance of different cross-supervision methods. In Fig. \ref{fig:cscs2}, the CPC* and CPS* perform multimodal response distillation by referring to supervision approaches of the CPC \cite{Ke2020CPC} and CPS \cite{Chen2021CPS}, respectively. Compared to the above methods, our CSCS achieves the best performance.

    % CSCS    
    \begin{table*}[t]
    \centering
    \small
    \adjustbox{max width=1.0\textwidth}{
    \begin{tabular}{c|c|c|cccc|cccc} 
    \hline
    \multirow{2}{*}{Supervision} & \multirow{2}{*}{Annotation} & \multirow{2}{*}{Mode} & \multicolumn{4}{c|}{3D}                                           & \multicolumn{4}{c}{2D}                                             \\ 
    \cline{4-11}
                                 &                             &                       & mAP            & Veh.           & Ped.           & Cyc.           & mAP            & Veh.           & Ped.           & Cyc.            \\ 
    \hline
    \multirow{4}{*}{Weak}        & \multirow{4}{*}{2D Box}     & $\mbox{-}$            & 45.76          & 59.95          & 44.68          & \textbf{32.66} & 36.77          & 50.36          & 28.41          & \textbf{31.53}  \\ 
    \cline{3-11}
                                 &                             & CPC* \shortcite{Ke2020CPC}        & 45.77          & 61.59          & 44.20          & 31.52          & 25.49          & 47.99          & 20.23          & 8.25            \\ 
    \cline{3-11}
                                 &                             & CPS* \shortcite{Chen2021CPS}      & 46.12          & 60.62          & 44.75          & 32.98          & 36.30          & 50.49          & \textbf{31.31} & 27.09           \\ 
    \cline{3-11}
                                 &                             & CSCS (ours)           & \textbf{46.93} & \textbf{62.46} & \textbf{45.73} & 32.59          & \textbf{37.20} & \textbf{51.13} & 29.05          & 31.42           \\
    \hline
    \end{tabular}
        }
    \caption{Comparisons of different cross-supervision methods.}
    \label{tab:cscs1}
    \end{table*}

    \noindent\textbf{Scaling up Weakly Supervised Training.} In order to further verify the effectiveness of our weakly supervised method, we compare the performance of our method on the expanded dataset in Tab. \ref{tab:scaling up training}. On the training set, there are only 23,691 frames with 3D mask annotations, while the entire dataset contains 158,081 frames with 2D box annotations. This is approximately 6.67 times more data than the fully supervised dataset. Based on the weak supervision of 2D boxes, our method achieves 98.22\% of the fully supervised performance on the semantic segmentation task. Furthermore, it surpasses the performance of full supervision in the cyclist class. Additionally, when combining our method with 3D boxes, the performance exceeds that of full supervision by 0.52\% mAP and 1.26\% mIoU.
    
    % Scaling up Weakly supervised Training
    \begin{table*}[ht]
    \centering
    \small
    \adjustbox{max width=1.0\textwidth}{
    \begin{tabular}{c|c|c|c|c|cccc|cccc} 
    \hline
    \multirow{2}{*}{Supervision} & \multirow{2}{*}{Annotation} & \multirow{2}{*}{Model} & \multirow{2}{*}{Data} & \multirow{2}{*}{Ours} & \multicolumn{4}{c|}{AP}                        & \multicolumn{4}{c}{IoU}                         \\ 
    \cline{6-13}
                                 &                             &                        &                       &                       & mAP           & Veh.  & Ped.  & Cyc.           & mIoU          & Veh.  & Ped.  & Cyc.            \\ 
    \hline
    Full                         & 3D Mask                     &                        & 23691                 & $\mbox{-}$            & 57.59         & 80.30 & 57.18 & 35.29          & 80.22         & 96.67 & 82.47 & 61.51           \\ 
    \cline{1-2}\cline{4-13}
    \multirow{5}{*}{Weak}        & \multirow{3}{*}{3D Box}     &                        & 23691                 & $\mbox{-}$            & 48.29         & 68.95 & 45.94 & 29.99          & 72.61         & 89.68 & 73.58 & 54.58           \\ 
    \cline{4-13}
                                 &                             & SparseUNet             & 158081                & $\mbox{-}$            & 53.19         & 74.35 & 50.84 & 40.79          & 77.63         & 91.40 & 78.25 & 63.24           \\ 
    \cline{4-13}
                                 &                             & \shortcite{shi2020parta2}                 & 158081                & $\checkmark$          & \underline{58.11} & 79.23 & 54.00 & \underline{41.10}  & \underline{81.48} & 94.98 & 80.23 & \underline{69.24}   \\ 
    \cline{2-2}\cline{4-13}
                                 & \multirow{2}{*}{2D Box}     &                        & 23691                 & $\checkmark$          & 45.76         & 59.95 & 44.68 & 32.66          & 76.30         & 89.50 & 78.94 & 60.44           \\ 
    \cline{4-13}
                                 &                             &                        & 158081                & $\checkmark$          & 49.07         & 62.09 & 48.86 & \textbf{36.27} & 78.79         & 89.94 & 82.31 & \textbf{64.12}  \\
    \hline
    \end{tabular}
        }
    \caption{Scaling up weakly supervised training on 3D instance and semantic segmentation.}
    \label{tab:scaling up training}
    \end{table*}

\end{document}